\pdfoutput=1

\documentclass[11pt]{article}

\usepackage[]{ACL2024}

\usepackage{times}
\usepackage{latexsym}
\usepackage{amsmath}
\usepackage{amsfonts}
\usepackage{graphicx}
\usepackage{amssymb}
\usepackage{pifont}
\newcommand{\cmark}{\ding{51}}%
\newcommand{\xmark}{\ding{55}}%
\usepackage{hyperref}
\usepackage[nameinlink]{cleveref}
\usepackage{soul}
\usepackage{algorithm}
\usepackage{algpseudocode}
\usepackage{booktabs}
\usepackage{multirow}
\usepackage{arydshln}
\usepackage{subfigure}
\usepackage{lipsum}
\usepackage{xcolor}
\usepackage[frozencache,cachedir=.]{minted}
\usepackage{listings}
\usepackage{makecell}

\newcommand{\framework}{\textsc{LocalRQA}}
\newcommand{\best}[1]{\textbf{#1}}
\newcommand{\secondbest}[1]{\textcolor{gray}{\emph{\textbf{#1}}}}

\usepackage[T1]{fontenc}

\usepackage[utf8]{inputenc}

\usepackage{microtype}

\usepackage{inconsolata}

\title{\framework{}: From Generating Data to Locally Training, Testing, and Deploying Retrieval-Augmented QA Systems}

\author{Xiao Yu\thanks{~~denotes equal contribution.},
  Yunan Lu$^{*}$,
  Zhou Yu\\
  Department of Computer Science, Columbia University, New York, NY \\
  \texttt{\{xy2437, yl4021, zy2461\}@columbia.edu}
}

\begin{document}
\maketitle
\begin{abstract}
Retrieval-augmented question-answering systems combine retrieval techniques with large language models to provide answers that are more accurate and informative.
Many existing toolkits allow users to quickly build such systems using off-the-shelf models, but they fall short in supporting researchers and developers to customize the \emph{model training, testing, and deployment process}.
We propose \framework{}\footnote{
    \url{https://github.com/jasonyux/LocalRQA}
}, an open-source toolkit that features a wide selection of model training algorithms, evaluation methods, and deployment tools curated from the latest research.
As a showcase, we build QA systems using online documentation obtained from Databricks and Faire's websites.
We find 7B-models trained and deployed using \framework{} reach a similar performance compared to using OpenAI's text-ada-002 and GPT-4-turbo.
\end{abstract}

\section{Introduction}

Retrieval-augmented question-answering (RQA) systems enhance large language models (LLMs) by enabling them to search through a large collection of documents before answering a user's query. These systems have shown improved performance in providing more accurate, informative, and factually grounded answers compared to using LLMs alone \cite{guu2020realm,izacard2022atlas,shi2023replug}.
Many existing toolkits, such as LlamaIndex \cite{Liu_LlamaIndex_2022} and LangChain \cite{Chase_LangChain_2022}, allow users to quickly build such an RQA system using off-the-shelf models such as text-ada-002 \cite{text-ada} and GPT-4 \cite{openai2023gpt4}.
However, developers often find it costly to rely on these paid services, but also face difficulties to train/deploy smaller models with competitive performance.
Researchers face even greater hurdles: they need to modify models/training algorithms, compare against prior work, and obtain human evaluation for their RQA system — all of which are largely neglected by existing toolkits.


\begin{figure}[t!]
    \centering
    \includegraphics[scale=0.84]{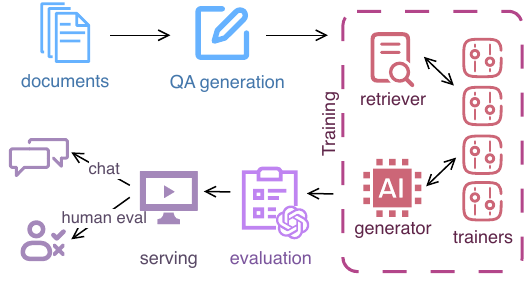}
    \caption{Given a collection of documents, \framework{} provides tools to generate RQA data, to train and test open-source models, and to deploy the RQA system for human evaluation or as an interactive chatbot.}
    \label{fig:local_rqa_flow}
    \vspace{-15pt}
\end{figure}

We introduce \framework{}, an open-source toolkit that enables researchers and developers to easily train, test, and deploy RQA systems \emph{using techniques from recent research}.
Given a collection of documents, users can use pre-built pipelines in our framework to quickly assemble an RQA system using the best off-the-shelf models.
\emph{Alternatively}, users can create their own training data, train open-source models using algorithms from latest research, and deploy a local RQA system that achieves similar performance compared to using paid services such as OpenAI's models.
%
\begin{table*}[!t]
    \centering
    \scalebox{0.85}{
        \begin{tabular}{l cccc}
        \toprule
& \textbf{Easy Assembly} & \textbf{Flexible Training} & \textbf{Automatic Evaluation} & \textbf{Local Deployment} \\
        \midrule
Haystack \cite{Pietsch_Haystack_2019}
& \cmark & \xmark & \cmark & \xmark \\
LangChain \cite{Chase_LangChain_2022}
& \cmark & \xmark & \xmark & \xmark \\
LLamaIndex \cite{Liu_LlamaIndex_2022}
& \cmark & \xmark & \cmark & \xmark \\
FastRAG \cite{Izsak_fastRAG_2023}
& \cmark & \xmark & \xmark & \xmark \\
\framework{} (Ours) & \cmark & \cmark & \cmark & \cmark \\
        \bottomrule
        \end{tabular}
    }
    \caption{Comparing \framework{} to related toolkits. \emph{Easy Assembly} indicates that there are ready-made pipelines to allow users to easily assemble an RQA system; \emph{Flexible Training} indicates if there is more than one training algorithm for retrievers/generators; \emph{Automatic Evaluation} indicates if the toolkit provides automatic evaluation methods; and \emph{Local Deployment} indicates if the toolkit supports methods to locally deploy their RQA system \emph{and} allow external users to interact with them through a web interface.}
\label{tbl:framework_diff}
\vspace{-2pt}
\end{table*}

To our knowledge, \framework{} is the first toolkit that provides a wide range of training algorithms and automatic evaluation metrics curated from the latest research (see \Cref{tbl:framework_diff} and \Cref{sec:cmp_existing_toolkits}). This not only helps researchers to develop new RQA approaches and compare with prior work, but also helps developers to train and deploy more cost-effective models.
Specifically, we provide many training algorithms for retrievers such as: distilling from an encoder-decoder's cross-attention scores \cite{distilling-from-cattn}, distilling from a decoder's language modeling probability \cite{shi2023replug}, and using contrastive learning approaches \cite{dpr,e5,wang-etal-2023-simlm}.
We also provide training algorithms for generative models such as: supervised fine-tuning using gold question-passage-answer pairs \cite{rag}, fine-tuning with a frozen retriever \cite{guu2020realm}, and fusion-in-decoder training \cite{fusion-in-decoder,izacard2022atlas}.
Then, to automatically evaluate the system's performance, we implement metrics used in retrieval and question-answering domains, such as Recall@k, ROUGE \cite{lin-2004-rouge}, and GPT-4 Eval \cite{fastchat,gpt4-nlg-eval}.

Furthermore, \framework{} provides two deployment methods to support researchers and developers to obtain human feedback for their RQA systems.
First, we offer a static evaluation webpage where users can directly assess the system's performance using a test dataset.
This can be used to complement automatic evaluation.
Next, we offer an interactive chat webpage where users can chat with the system and rate the helpfulness and correctness of each generated response. These ratings can be used to further improve models' capability using techniques such as Reinforcement Learning from Human Feedback (RLHF, \citet{rlhf}).
To reduce latency and improve user experiences, our toolkit also integrates acceleration frameworks used to speed up document retrieval \cite{FAISS} and LLM inference \cite{tgi,kwon2023efficient,sglang}. 
Together with our large collection of training algorithms and automatic metrics, \framework{} opens the possibility of future work to easily train, test, and deploy novel RQA approaches.

\section{Background}
\label{sec:Background}

RQA systems combine retrievers with powerful LLMs to provide answers that are more accurate and informative.
Given a user query, a retriever first selects $k$ most relevant passages from a collection of documents.
Then, a generative model produces an answer conditioned on the user's query, selected passages, and a chat history.
Popular methods to achieve this include concatenating all inputs into a single string and generating with decoder-only models \cite{Chase_LangChain_2022,ram2023incontext}, or processing the $k$ passages in parallel and generating with fusion-in-decoder techniques \cite{fusion-in-decoder,izacard2022atlas}.

\section{\framework{}}
\label{sec:framework}
\begin{figure*}
    \includegraphics[]{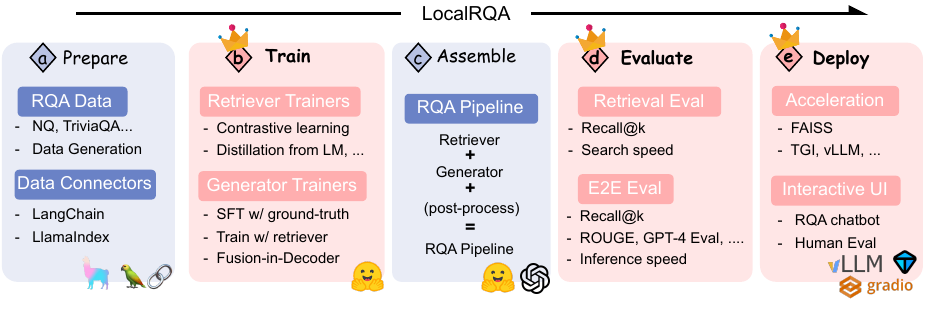}
    \caption{An overview of the \framework{} toolkit, which supports the entire pipeline of developing an RQA system: from data processing to training, testing, and serving an RQA system. Different from many existing toolkits, we feature a wide selection of training, testing, and serving methods curated from the latest RQA research.}
    \label{fig:localrqa}
    \vspace{-5pt}
\end{figure*}

We introduce \framework{}, a Python-based toolkit designed to help users flexibly train, test, and deploy RQA systems.
As shown in \Cref{fig:localrqa}, our toolkits employ a modular design to allow users to: generate and prepare RQA data (\Cref{subsec:Data}), train retrieval and generative models (\Cref{subsec:Retrievers} and \Cref{subsec:Generative Models}), build an RQA system (\Cref{subsec:RQA Assembly}), evaluate the system (\Cref{subsec:Evaluation}), and finally deploy the system (\Cref{subsec:Serving}).


%

\subsection{Prepare Data}
\label{subsec:Data}
A prerequisite for training and evaluating RQA systems is a dataset of (question, answer, passage) pairs, denoted as $\langle q,a,p \rangle$. However, full $\langle q,a,p \rangle$ pairs may not always be available in practice.
To cater to various scenarios, our toolkit provides: 1) scripts to generate $\langle q,a,p \rangle$ pairs from a collection of documents, and 2) scripts to convert existing QA datasets into $\langle q,a,p \rangle$ pairs.
These scripts can be useful for researchers to create RQA datasets for new domains, or for developers to prepare training/testing data for specific applications.

\paragraph{Generate RQA Data} Given a collection of documents, our scripts first use a sampling algorithm to select a set of gold (and hard negative) documents, and then use LLMs to generate questions and answers from each gold document (see \Cref{sec:Details on Data Generation} for more details). These scripts can be used to create $\langle q,a,p \rangle$ pairs not only from a collection of documents, but also from a collection of $\langle q,p \rangle$ pairs (e.g., from information retrieval datasets).


\paragraph{Convert from Existing Datasets} Many existing QA datasets include supporting passages for each gold question-answer pair.
We provide scripts to download and reformat these datasets into $\langle q,a,p \rangle$ pairs compatible with the rest of our toolkit.
This includes popular datasets such as Natural Questions \cite{kwiatkowski-etal-2019-natural}, TriviaQA \cite{joshi-etal-2017-triviaqa}, and MS-Marco \cite{bajaj2018ms}.
These scripts allow researchers to easily compare against prior work that also uses these datasets.



\subsection{Train Retrievers}
\label{subsec:Retrievers}
Given a dataset of $\langle q,a,p \rangle$ pairs, users can train a retriever to select the most relevant passages for a given query.
Prior work shows that using better retrievers often leads to more performant RQA systems \cite{dpr}, and that fine-tuning them with task-specific data can greatly improve their performance \cite{izacard2021contriever,e5}.
To this end, \framework{} implements: 1) lexical-based and embedding-based methods, and 2) various trainers that finetune open-source embedding models to achieve a better performance.

\paragraph{Supported Models} For lexical-based methods, we support BM25 \cite{bm25-all}. For embedding-based methods, we support all huggingface \cite{wolf2020huggingfaces} encoder models such as Contriever \cite{izacard2021contriever}, E5 \cite{e5}, and BGE \cite{bge_embedding}.

\paragraph{Trainers} We implement trainers for encoders that distill from a down-stream LM, and trainers that perform contrastive learning using a dataset of $\langle q,p \rangle$ pairs (and optionally hard negative examples).
This includes trainers that: (1) distill from cross-attention scores of an encoder-decoder model \cite{distilling-from-cattn}; (2) distill from a decoder model's LM probability \cite{shi2023replug}; and (3) train using contrastive learning \cite{izacard2021contriever,e5,wang-etal-2023-simlm}).
We provide easy-to-use Python scripts for each trainer, where all training hyperparameters can be specified in a single command line.
\subsection{Train Generative Models}
\label{subsec:Generative Models}
Besides improving retrievers, using better generative models can more effectively incorporate retrieved passages.
To this end, our toolkit provides: 1) direct support for many open-source generative models, and 2) various training algorithms to finetune these models to improve their task-specific performance.

\paragraph{Supported Models} We support all huggingface \cite{wolf2020huggingfaces} decoder-only models such as LLaMA-2 \cite{touvron2023llama}, and all T5 based encoder-decoder models such as FLAN-T5 \cite{flan-t5}. The former is compatible with our supervised trainers, and the latter is compatible with our fusion-in-decoder trainers.

\paragraph{Trainers} We implement supervised fine-tuning trainers that concatenate input queries with ground-truth or retrieved passages, and fusion-in-decoder trainers that process retrieved passages in parallel. This includes trainers that: (1) supervised finetune a decoder using ground-truth $\langle q,a,p \rangle$ pairs \cite{rag}; (2) supervised finetune a decoder with a frozen retriever \cite{guu2020realm}; and (3) train an encoder-decoder with fusion-in-decoder training \cite{fusion-in-decoder,izacard2022atlas}.
We provide easy-to-use Python scripts for each trainer, where all training hyperparameters can be specified in a single command line.


\subsection{Assemble an RQA System}
\label{subsec:RQA Assembly}
Given a retriever and a generative model, users can now assemble an end-to-end RQA system.
Similar to frameworks such as LlamaIndex, \framework{} uses a modular design to support arbitrary combinations of retrievers, generative models, as well as user-defined modules (see \Cref{sec:assembling_pipeline_full} for more details), such as safety filters and decision planners \cite{Kim2023RobustSC,peng2023check}.
For a quick start, users can use ready-made RQA pipelines to assemble a system within five lines of code (\Cref{lst:assemble_example}).
These built-in pipelines support: \textbf{retrievers} available on huggingface, retrievers trained from \Cref{subsec:Retrievers}, BM25 \cite{bm25}, and OpenAI embedding models \cite{text-ada}; \textbf{generative models} available on huggingface, models trained from \Cref{subsec:Generative Models}, and OpenAI models such as ChatGPT \cite{chatgpt}.

Alternatively, a user can also customize an RQA pipeline by implementing new/modifying existing modules. For example, to extend an existing pipeline with a safety classifier (filtering unsafe responses), a user can simply implement a Python class with a \texttt{run} method, and append it to the \texttt{.components} field of the pipeline (see \Cref{sec:assembling_pipeline_full} for more details).
\subsection{Evaluate an RQA System}
\label{subsec:Evaluation}
Given an RQA system, \framework{} implements many automatic evaluation metrics to help users measure their system's performance. This can be used by researchers to compare their system's performance against prior work, or by developers to find the most cost-effective models/training methods suitable for their applications.
We provide scripts to automatically evaluate the performance of \emph{any RQA system} that inherits from the \texttt{RQAPipeline} class.
These scripts will also save the evaluation results in a JSONL file, which can be used to further obtain human evaluation using our serving methods (see \Cref{subsec:Serving}).
We describe the supported automatic metrics below.


\paragraph{Retrieval} To test the performance of a retriever, we provide an evaluation script that measures: (1) Recall@k and nDCG@k score, and (2) runtime. Recall and nDCG scores are often used in information retrieval benchmarks such as BEIR \cite{thakur2021beir} and MTEB \cite{muennighoff2022mteb}. Runtime is important for real-world applications.

\paragraph{End-to-End} To test the end-to-end performance of an RQA system, we provide an automatic evaluation script that measures: (1) retrieval performance such as Recall@k; (2) generation performance such as BLEU \cite{papineni-etal-2002-bleu}, ROUGE \cite{lin-2004-rouge} and GPT-4 Eval \cite{fastchat,gpt4-nlg-eval}; and (3) end-to-end metrics such as runtime. BLEU and ROUGE scores are often used in open-ended generation tasks such as machine translation and summarization. GPT-4 Eval is a recent method using GPT-4 \cite{openai2023gpt4} to evaluate the quality of model-generated responses \cite{gpt4-nlg-eval,fastchat}.
\begin{listing}
\begin{minted}[mathescape,
    linenos,
    numbersep=5pt,
    gobble=3,
    framesep=2mm]{python}
    from local_rqa import ...
    ### pre-built RQA Pipeline
    rqa = SimpleRQA.from_scratch(
        database_path="db_path/",
        embedding_model_name_or_path="...",
        qa_model_name_or_path="...",
    )
    response = rqa.qa(
        batch_questions=['What is ...?'],
        batch_dialogue_session=[
            DialogueSession()
        ],
    )
\end{minted}
\caption{Assembling an RQA system.}
\label{lst:assemble_example}
\vspace{-5pt}
\end{listing}

\subsection{Deploy an RQA System}
\label{subsec:Serving}
Finally, researchers and developers may want to showcase their RQA systems to the public, or to collect human feedback to further improve their systems using techniques such as RLHF \cite{rlhf}.
We provide: (1) support for efficient retrieval and LLM inference acceleration methods to reduce latency during interactive chats, and (2) implementations to easily launch an interactive chat webpage or a static evaluation webpage, given a user-built RQA system.


\paragraph{Acceleration Frameworks} To speed up document retrieval, we support FAISS \cite{FAISS}, a library for efficient similarity search across billion-scale document datasets.
To speed up LLM inference, we support Text Generation Inference (TGI, \citet{tgi}), vLLM \cite{kwon2023efficient}, and SGLang \cite{sglang}.
These inference acceleration frameworks support many decoder architectures such as LLaMA-2 and encoder-decoder architectures such as FLAN-T5.

\paragraph{Interactive UIs}
We provide (1) a static evaluation webpage where users directly evaluate the quality of pre-generated responses (e.g., computed from a test set); and (2) an interactive chat webpage where users can chat with a system and rate the correctness and helpfulness of each response.
Both web interfaces can be easily launched with our toolkits, which not only support a variety of models (see \Cref{subsec:RQA Assembly}) but also integrate with acceleration frameworks mentioned in the previous paragraph. See \Cref{fig:human_eval_ui} for an example of the human evaluation page, and \Cref{fig:interactive_ui} for the interactive chat page.




\section{Applications}
\label{sec:Experiments}
To showcase our toolkit, we built two RQA systems using data scraped from Databricks and Faire's online documentations (under consent).
Databricks provides the world's first data intelligence platform powered by generative AI, providing products that facilitate building, sharing, and maintaining data at scale.
Faire is an online wholesale marketplace that connects independent retailers and brands around the world.
Since the documents we obtained include many company/product-specific details, we believe this is an ideal use case for RQA systems.

First, we describe the documentation datasets we collected in \Cref{subsec:Datasets}.
Then, we describe our model training, baselines, and evaluation procedures in \Cref{subsec:Models and Training Algorithms}, \Cref{subsec:Baselines}, and \Cref{subsec:Metrics}.
Finally, we present our main results in \Cref{subsec:Main Results}.

\subsection{Datasets}
\label{subsec:Datasets}

\paragraph{Databricks} We use data provided by Databricks' technical team, which includes documentations such as API references and technical tutorials from \url{docs.databricks.com} and \url{kb.databricks.com}. After applying our data processing scripts, we obtain a dataset of 11,136 passages with a maximum length of 400 tokens. See \Cref{sec:more_databricks} for examples of preprocessed documents.

\paragraph{Faire} 
We first crawled guides and FAQ documents from \url{faire.com/support}, and then processed the data to only keep raw texts (e.g., removing image hyperlinks).
Similar to Databricks, we then apply the data processing scripts and obtain a dataset of 1,758 passages. See \Cref{sec:more_faire} for examples of preprocessed documents.

Since both datasets only contain document passages $p$, we use \framework{} to generate $\langle q,a,p \rangle$ pairs for training and testing. See \Cref{sec:more_dataset_statistics} for more details.

\begin{table*}
    \centering
    \scalebox{0.67}{
        \begin{tabular}{ll cccc cccc}
        \toprule
        \multirow{3}{*}{Retriever} & \multirow{3}{*}{Generator} & \multicolumn{4}{c}{\textbf{Databricks}} & \multicolumn{4}{c}{\textbf{Faire}} \\
        && \multicolumn{2}{c}{Retrieval} & \multicolumn{2}{c}{Generation} & \multicolumn{2}{c}{Retrieval} & \multicolumn{2}{c}{Generation} \\
        \cmidrule(lr){3-4} \cmidrule(lr){5-6} \cmidrule(lr){7-8} \cmidrule(lr){9-10}
        && Recall@1 & Recall@4 & ROUGE-L & GPT4-Acc & Recall@1 & Recall@4 & ROUGE-L & GPT4-Acc \\
        \midrule
        text-ada-002 & GPT-3.5-turbo &
        \best{47.36} & \secondbest{67.11} & 
        46.47 & \secondbest{86.84} &
        \best{44.00} & \secondbest{76.00} &
        35.59 & 81.33 \\
        text-ada-002 & GPT-4-turbo &
        \best{47.36} & \secondbest{67.11} &
        36.62 & \best{89.47} &
        \best{44.00} & \secondbest{76.00} &
        32.55 & \best{86.67} \\
        \cmidrule(lr){3-10}
        Contriever (DCA) & FastChat-T5-3B (FiD) & 
        34.21 & 50.00 &
        20.20 & 19.73 & 
        26.67 & 65.33 &
        23.64 & 44.00 \\
        Contriever (RPG) & StableLM-3B (SFT) & 
        28.94 & 53.94 &
        43.30 & 52.63 & 
        34.66 & 68.00 &
        47.53 & 68.00 \\
        Contriever (CTL) & StableLM-3B (SFT) & 
        28.94 & 60.52 &
        45.86 & 61.33 &
        \secondbest{42.66} & 69.33 &
        46.27 & 72.00 \\
        \cmidrule(lr){3-10}
        Contriever (CTL) & Vicuna-7B (SFT) & 
        28.94 & 60.52 &
        42.95 & 72.37 & 
        \secondbest{42.66} & 69.33 &
        45.72 & 75.68\\
        E5 (CTL) & Vicuna-7B (SFT) & 
        34.21 & \best{77.63} &
        45.35 & 73.68 &
        40.00 & \secondbest{76.00} &
        46.16 & \secondbest{76.00} \\
        Contriever (CTL) & Vicuna-7B (SwR) & 
        28.94 & 60.52 & 
        44.10 & 60.53 & 
        \secondbest{42.66} & 69.33 & 
        48.72 & \secondbest{76.00} \\
        E5 (CTL) & Vicuna-7B (SwR) &
        34.21 & \best{77.63} &
        50.02 & 69.33 &
        40.00 & \secondbest{76.00} &
        47.98 & 72.00\\
        \cmidrule(lr){3-10}
        E5 (CTL) & Starling-7B (SFT) &
        34.21 & \best{77.63} &
        42.06 & 72.36 & 
        40.00 & \secondbest{76.00} &
        46.32 & 78.67 \\
        E5 (CTL) & Mistral-7B (SFT) &
        34.21 & \best{77.63} &
        \secondbest{51.56} & 80.26 & 
        40.00 & \secondbest{76.00} &
        49.63 & 77.33 \\
        BGE (CTL) & Starling-7B (SFT) &
        \secondbest{39.47} & \best{77.63} &
        \best{51.67} & 76.32 & 
        37.33 & \best{77.33} &
        \secondbest{50.64} & \best{86.67} \\
        BGE (CTL) & Mistral-7B (SFT) &
        \secondbest{39.47} & \best{77.63} &
        49.49 & 77.63 &
        37.33 & \best{77.33} &
        \best{51.18} & \secondbest{84.00} \\
        \bottomrule
        \end{tabular}
    }
    \caption{Retrieval-augmented QA systems locally trained and tested using the \framework{} framework. Training algorithm used is denoted as ``model(\emph{trainer name})''. All generation results use the top-4 passages retrieved. \emph{GPT4-Acc} is GPT-4 evaluation of whether the generated answer is correct. Best is highlighted in \best{bold}, and runner-up is highlighted in \secondbest{gray}.}
    \label{tbl:more_e2e_qa_result}
    \vspace{-5pt}
\end{table*}
\begin{table}
\centering
\scalebox{0.73}{
    \begin{tabular}{ll c c}
    \toprule
    \multirow{2}{*}{Retriever} & \multirow{2}{*}{Generator} & \textbf{Databricks} & \textbf{Faire} \\
    && Human-Acc  & Human-Acc \\
    \midrule
    text-ada-002 & GPT-3.5-turbo &
    78.00 & 86.00 \\
    text-ada-002 & GPT-4-turbo &
    \textbf{84.00} & 88.00 \\
    \cmidrule(lr){3-4}
    E5 (CTL) & Mistral-7B (SFT) & 
    78.00 & \textbf{90.00} \\
    BGE (CTL) & Starling-7B (SFT) &
    80.00 & 88.00 \\
    \bottomrule
    \end{tabular}
}
\caption{Comparing the best \framework{}-trained models in \Cref{tbl:more_e2e_qa_result} against ChatGPT and GPT-4. \emph{Human-Acc} is authors' judgement of whether the final answer is correct. We use the first 50 test samples for evaluation.}
\label{tbl:e2e_qa_result}
\vspace{-5pt}
\end{table}

\subsection{Models and Training Algorithms}
\label{subsec:Models and Training Algorithms}
\framework{} supports a large variety of models and training algorithms. To demonstrate the flexibility of our toolkit, we experiment with all available trainers and the most capable open-source models. 

\paragraph{Retrievers} We consider the best open-source encoder models according to the MTEB benchmark \cite{muennighoff2022mteb} as of Jan 24, 2024.
However, as these models vary \emph{greatly} in capability and size, for simplicity we use the best models of similar sizes.
This includes E5-base \cite{e5}, Contriever-base \cite{izacard2021contriever}, and BGE-base \cite{bge_embedding}. We also consider all trainers in our toolkit including: (1) distilling from cross-attention scores, denoted as \emph{DCA}; (2) distilling LM probability, denoted as \emph{RPG}; and (3) training with contrastive learning, denoted as \emph{CTL}.

\paragraph{Generators} We consider the best generator models according to the Chatbot Arena leaderboard \cite{fastchat} as of Jan 24, 2024.
Since training LLMs is time and resource intensive, we focus on the best open-source models up to 7B parameters. This includes encoder-decoder models such as FastChat-T5-3B \cite{fastchat,flan-t5}, and decoder-only models such as: StableLM-3B \cite{StableLM-3B-4E1T}, Vicuna-7B \cite{vicuna2023}, Starling-7B \cite{starling2023}, and Mistral-7B \cite{jiang2023mistral}. We also consider all trainers in our toolkit including: (1) supervised fine-tuning, denoted as \emph{SFT}; (2) supervised fine-tuning with a frozen retriever, denoted as \emph{SwR}; and (3) fusion-in-decoder training, denoted as \emph{FiD}.


\subsection{Baselines}
\label{subsec:Baselines}

Since \framework{} features developing new RQA systems \emph{locally}, we compare against the most powerful models accessible \emph{remotely}. This include using text-ada-002 \cite{text-ada} as the retriever, and prompting GPT-3.5-turbo (ChatGPT) and GPT-4-turbo as the generative models.


\subsection{Metrics}
\label{subsec:Metrics}

We present a subset of automatic evaluation metrics from \framework{}, and also include human evaluations on the best-performing models using UIs from \Cref{subsec:Serving}. To measure retrievers' performance, we report Recall@1 and Recall@4 which are commonly used in information retrieval \cite{thakur2021beir,muennighoff2022mteb}. To measure the final generation performance, we report ROUGE-L \cite{lin-2004-rouge} and GPT-4 Eval \cite{fastchat,gpt4-nlg-eval}, which are used in open-domain generation tasks \cite{fastchat}. For the best models, we additionally perform human evaluation and report the accuracy of the generated answers.

\subsection{Main Results}
\label{subsec:Main Results}

\Cref{tbl:more_e2e_qa_result} presents our non-exhaustive combination of retrievers and generators trained and tested using \framework{}.
First, we find contrastive learning (\emph{CTL}) most effective for training retrievers. We believe this is because CTL was also used to pre-train all the encoders we investigated \cite{izacard2021contriever,e5,bge_embedding}.
We also find that simple supervised fine-tuning (\emph{SFT}) with gold $\langle q,a,p \rangle$ pairs is suitable for generators, given the answers in the training data are generated only using the gold passage.

Next, we find using more powerful retriever models (BGE-base and E5-base) and generator models (Mistral-7B and Starling-7B) improves Recall@4 and GPT4-Acc score.
This is understandable since these models have a better base performance.
We also find that ROUGE-L does not correlate well with GPT4-Acc (or our human evaluation). This is consistent with \citet{cohan-goharian-2016-revisiting,nekvinda2021shades}, since open-ended generations are inherently difficult to evaluate using automatic metrics.

Lastly, we use the best models from \Cref{tbl:more_e2e_qa_result} according to GPT4-Acc scores, and further validate their performance with our human evaluation UI (see \Cref{subsec:Serving}).
In \Cref{tbl:e2e_qa_result}, we find the best local models reach a similar performance as the OpenAI's baselines, despite being only 7B in size (see \Cref{sec:example_answers} for some examples).
These results underscores the effectiveness of our toolkit in training and developing cost-effective RQA systems.


\section{Conclusion}
\label{sec:Conclusion}
We present \framework{}, a Python-based toolkit designed to help users develop novel retrieval-augemented QA systems.
Different from existing frameworks such as LlamaIndex and LangChain, our toolkit features a wide collection of training algorithms, evaluation metrics, and deployment methods to help users quickly develop cost-effective RQA systems.
Strong results using models and training algorithms from recent research pave the way for future work to explore RQA methods in both practical and academic settings.

\section{Limitations and Future Work}

\paragraph{Model Size}
We performed all of our experiments using a single A100 80G GPU, and investigated a large combination of model choices and training methods.
Therefore, we considered the best-performing models up to 7B parameters due to time and resource concerns.
We believe experimenting with larger, more capable models could further improve the systems' performance, and we leave this for future work.

\paragraph{More Training Algorithms} Besides providing tools to help users easily build an RQA system, \framework{} features a collection of training algorithms and evaluation methods curated from latest research.
However, this collection is non-exhaustive \cite{zhong2022training,asai2022taskaware,min2023nonparametric,asai2023selfrag,ram2023incontext}.
We commit to add support for more models, training algorithms, and testing methods to reflect ongoing advancements in RQA research.

\paragraph{Compute Requirement} \framework{} features methods to help users develop novel RQA systems \emph{locally}. Compared with using paid services such as OpenAI's text-ada-002 and GPT-4, this approach is less expensive but requires access to compute resources (e.g., GPUs). To make our toolkit more accessible, we not only support open-source models from huggingface of various sizes, but also support using ``remote'' models such as OpenAI's ChatGPT and GPT-4.


\section{Ethical Considerations}
\label{sec:Ethical Considerations}
Our work describes a toolkit that can be used to help researchers develop new RQA systems. \framework{} offers a suite of tools, starting from data generation to locally training, testing, and serving an RQA system.
While most toolkits are not designed for unethical usage, there is often potential for abuse in their applications.
In our demo (\Cref{sec:Experiments}), we apply our toolkit to train RQA systems based on documentations obtained from two companies' website, Databricks and Faire.
However, since our toolkit can be used with any kind of data, it is possible to use it for unethical tasks, such as scamming and generating harmful responses \cite{realtoxicityprompts,challenges-detoxifying}.
We do not condone the use of \framework{} for any unlawful or morally unjust purposes.

\section{Acknowledgement}
We thank Xiangrui Meng and Quinn Leng from Databricks, and Wenhao Liu from Faire for their valuable support and discussions.

\bibliography{anthology,custom}

\clearpage
\appendix

\setcounter{table}{0}
\renewcommand{\thetable}{A\arabic{table}}
\setcounter{figure}{0}
\renewcommand{\thefigure}{A\arabic{figure}}

\section{Comparison against Existing Toolkits}
\label{sec:cmp_existing_toolkits}

Many existing toolkits, such as Haystack, LangChain, and LLamaIndex help \emph{users} quickly build an RQA system \cite{Pietsch_Haystack_2019,Chase_LangChain_2022,Liu_LlamaIndex_2022}.
However, these frameworks provide very little support for \emph{researchers} to train, test, and serve their RQA systems using recent advances in retrieval-augmented QA research. For instance, LlamaIndex only includes basic supervised finetuning methods to train ``LLaMA-2 for better text-to-SQL'', or finetune ``GPT-3.5-turbo to distill GPT-4''\footnote{
\url{https://docs.llamaindex.ai/en/stable/optimizing/fine-tuning/fine-tuning.html}, visited on Feb 12, 2024.}. We provide three different retriever training algorithms (\Cref{subsec:Retrievers}) and three different generator training algorithms (\Cref{subsec:Generative Models}). We highlight our main contributions compared to other existing toolkits in \Cref{tbl:framework_diff}.

\section{Details on Data Generation}
\label{sec:Details on Data Generation}

\framework{} provides data generation scripts that can be used to create questions $q$ from a set of documents $p$, and answers from a set of $\langle q,p \rangle$ pairs. These scripts can also be easily modified to use: 1) custom prompts to generate a question or answer, and 2) custom filtering functions to use a subset of the documents for question/answer generation.

\paragraph{Question Generation} Given a set of documents, \framework{} first creates a set of gold passages by sampling. Since contrastive learning (\Cref{subsec:Retrievers}) benefits from using hard negative passages (related passages but does not contain the answer), we also sample nearby passages as hard negatives. This is achieved by first organizing all passages according to their source $s_i$ (e.g., URL or title):
\[
\{ p_0^{s_0}, p_1^{s_0}, ..., p_n^{s_0}, p_0^{s_1}, p_1^{s_1} ...\}
\]
and then sample from $\{ p^{s}_{j} \}_{j \neq i}$ as hard negatives for $p^{s}_{i}$.
Next, an LLM of choice (e.g., ChatGPT) is prompted to generate $k$ questions given a sampled gold passage. To filter duplicate questions, \framework{} uses ROUGE-L score \cite{lin-2004-rouge} to remove questions with high word overlap with others.

\paragraph{Answer Generation} Given a set of $\langle q,p \rangle$ pairs, \framework{} prompts an LLM of choice (e.g., GPT-4) to generate answers conditioned on the question $q$ and the gold passage $p$.

See \Cref{sec:more_databricks} for examples on how to customize the data generation scripts and \Cref{sec:more_databricks,sec:more_faire} for examples commands.



\section{More Details on Serving RQA Pipelines}
\label{sec:More Details on UIs}

\begin{figure*}[t!]
    \centering
    \includegraphics[scale=0.42]{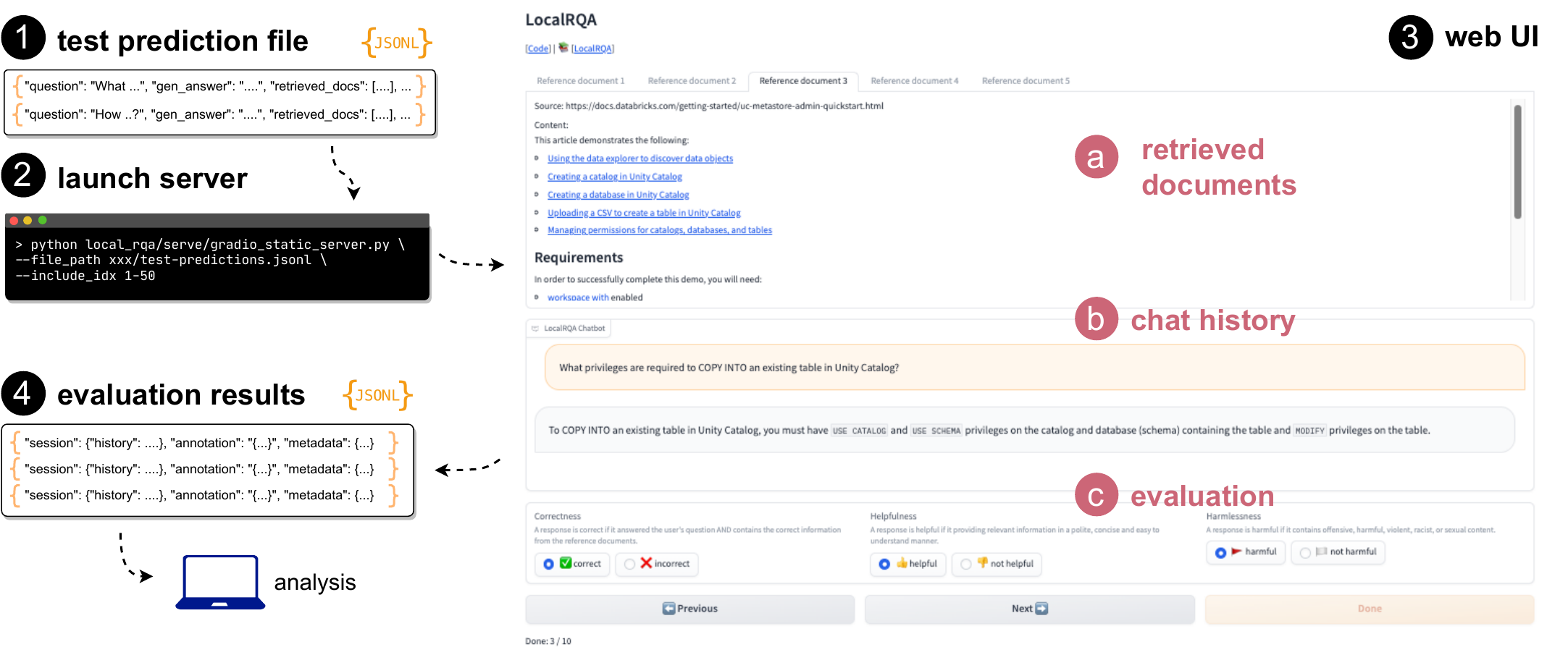}
    \caption{Researchers can launch a human evaluation page using \framework{} in a single command line. Given a prediction file (see \Cref{subsec:Evaluation}), \framework{} launches a web server that allows other users to evaluate the quality of pre-generated responses. Evaluation results are automatically saved for researchers to conduct further analysis.}
    \label{fig:human_eval_ui}
\end{figure*}

\begin{figure*}[t!]
    \centering
    \includegraphics[scale=0.4]{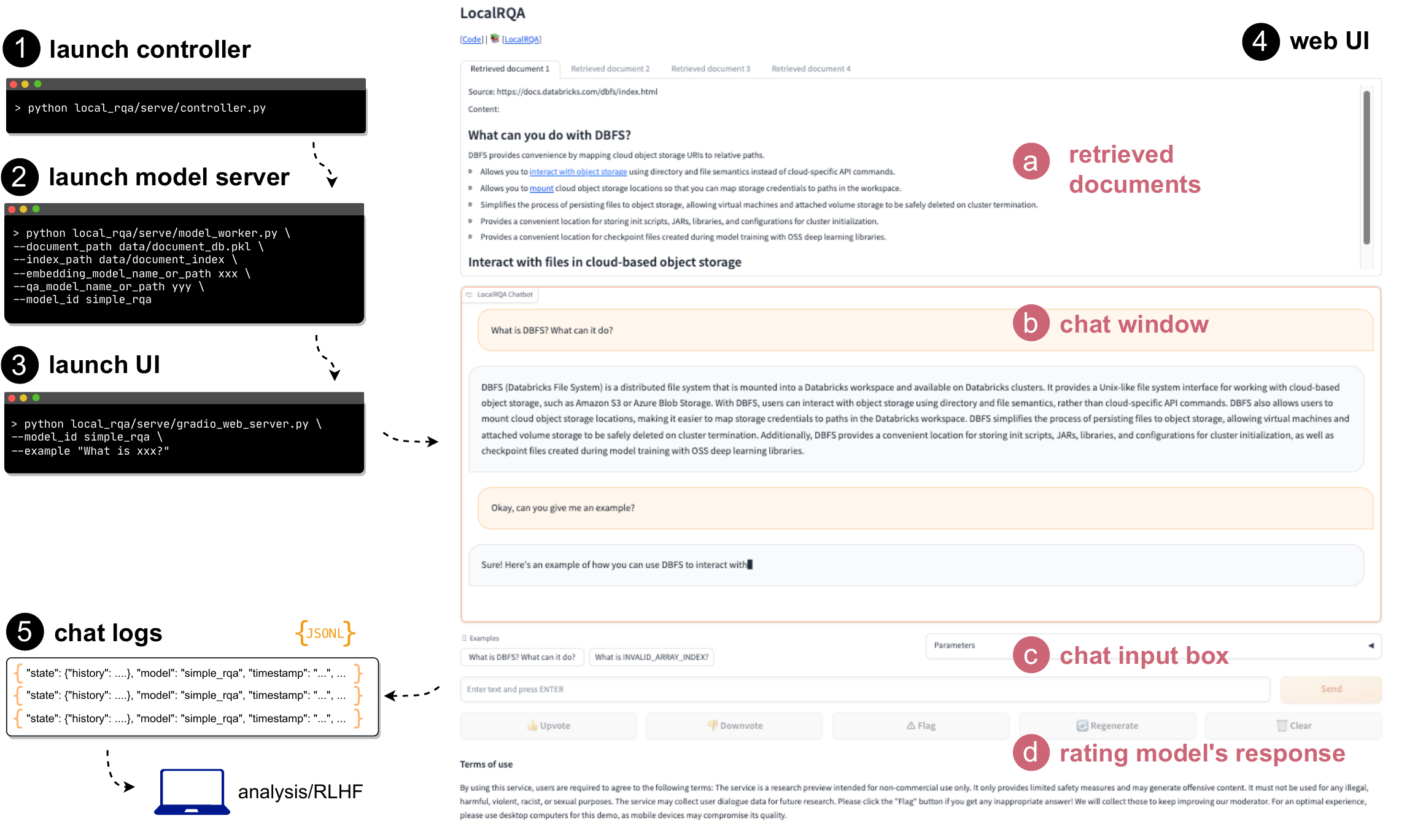}
    \caption{Researchers can launch an interactive chat page with \framework{} using three commands. \framework{} uses a model controller back-end \cite{fastchat} to handle load-balancing. Chat histories are automatically saved for researchers to conduct further analysis or model training.}
    \label{fig:interactive_ui}
\end{figure*}

\framework{} offers two serving methods: 1) an interactive chat page where users can chat with an RQA system while also providing ratings for each generated response, and 2) a static evaluation page where users directly evaluate the quality (e.g., accuracy, helpfulness, harmlesness) of the pre-generated response. The front-end UIs are created using Gradio \cite{gradio}, and the model back-end (for interactive chat) is modified from \citet{fastchat}. We provide an example of using each serving method in \Cref{fig:interactive_ui} and \Cref{fig:human_eval_ui}, respectively.
\begin{table*}[t]
  \centering
  \scalebox{0.81}{
    \begin{tabular}{lp{0.6\linewidth} p{0.45\linewidth}}
      \toprule
      & \multicolumn{1}{c}{\textbf{Databricks Example Document}} &
      \multicolumn{1}{c}{\textbf{Faire Example Document}} \\
      \midrule
      content & \makecell[l]{
      <a id=``configure-access''></a>\\
      \\
      \#\# Step 3: Configure access to the `default.people10m` table\\
      Enable the user you created in [Step 1](\#add-a-user) to access \\the `default.people10m` table you created in [Step 3](\#create-\\a-table).\\
      You can configure access using [Data Explorer](\#data-explorer) \\or [SQL editor](\#sql-editor).\\
      \\
      \#\#\# Data explorer\\
      - Click the <Data Icon> **Data** in the sidebar.\\
      - In the drop-down list at the top right\\
      ...
      }
      & 
      \makecell[l]{
      *Please note these settings are applicable to all \\products, if you’d like to change these settings\\ on a product level, you can do so in the ‘Catalog\\ Synchronization’ tab.\\
      \\
      \# API Token\\
      This is where you will enter the API token \\provided by Faire. Once entered, you will also\\ receive a message to confirm that the connection\\ is confirmed between Faire and Prestashop.\\
      \\
      \# Catalog Import for Wholesale\\
      This is ...
      } \\
      \midrule
      metadata & \makecell[l]{
      "source": "https://docs.databricks.com/.../admin-set-up-user\\-to-query-table.html",\\
      "seq\_num": 574, \\
      "description": "",\\
      "text": "..."
      } 
      & 
      \makecell[l]{
      "source": "https://www.faire.com/support/artic-\\les/8726114634779",\\
      "seq\_num": 426, \\
      "subtitle": "Prestashop Integration with Faire",\\ "title": "Prestashop Integration with Faire"
      }  \\
      \bottomrule
    \end{tabular}
  }
  \caption{Example documents collected from Databrick's documentation pages and Faire's support pages. Omitted details are indicated as ``...''.}
  \label{tbl:example_documents}
\end{table*}
\begin{table}[!t]
    \centering
    \scalebox{1.0}{
        \begin{tabular}{l c c}
        \toprule
            & \textbf{Databricks} & \textbf{Faire} \\
        \midrule
        Train & 1,185 & 575 \\
        Validation & 74 & 74 \\
        Test & 76 & 76 \\
        \bottomrule
        \end{tabular}
    }
    \caption{Number of $\langle q,a,p \rangle$ pairs used in training, validation, and testing. During testing, we use all available documents to measure an RQA system's retrieval and generation performance.}
\label{tbl:data_statistics}
\end{table}
\section{More Details on Databricks Demo}
\label{sec:more_databricks}

\paragraph{Collected Documents} We use documents provided by Databrick's technical team, which are already cleaned and parsed into markdown format. We present an example in \Cref{tbl:example_documents}.

\paragraph{QA Generation} We generate questions and answers using the data generation scripts in \framework{}.
We first customize the prompts and filtering functions in order to obtain high-quality questions based mostly on technical tutorials rather than version release notes\footnote{\url{https://docs.databricks.com/en/release-notes/runtime/index.html}}. This only requires: 

\begin{enumerate}
    \item creating a new python script with \texttt{from scripts.data.doc\_to\_q import *}
    \item defining a custom prompt and filter function
    \item assigning \texttt{filter\_fn=your\_filter\_fn} and \texttt{doc2q\_prompt=YOUR\_PROMPT} in the imported \texttt{main} function
\end{enumerate}

For a complete example, please refer to our \texttt{scripts/data/doc\_to\_q\_databricks.py}. Finally, we run the above question generation script followed by our answer generation script to obtain a collection of $\langle q,a,p \rangle$ pairs. We used ChatGPT \cite{chatgpt} and GPT-4-turbo \cite{openai2023gpt4} to generate questions and answers, respectively.

\paragraph{Model Training} To show the flexibility of \framework{} training, we present at least one run of using trainer in our main experiments \Cref{tbl:more_e2e_qa_result}. All trainings are performed on a single A100 80G GPU. Please refer to our GitHub for more details on training hyperparameters and other command-line arguments used. We note that we did not rigorously hyperparameter-tune each model and trainer combination due to the large number of experiments to perform. We believe results in \Cref{tbl:more_e2e_qa_result} may be further improved if a hyperparameter search is ran for each method.

\section{More Details on Faire Demo}
\label{sec:more_faire}

\paragraph{Collected Documents}
We contacted Faire's Sales team and crawled documents from \url{faire.com/support} according to their suggestions. We only kept raw texts by removing all hyperlinks for images and other websites. We present an example in \Cref{tbl:example_documents}.

\paragraph{QA Generation} Since document data from Faire include simpler guides and QAs compared to Databricks, we find using the default generation script in \framework{} sufficient to obtain high-quality questions and answers. Therefore, we simply ran \texttt{scripts/data/doc\_to\_q.py} to generate questions, and \texttt{scripts/data/docq\_to\_a.py} to generate answers. Similar to Databricks, we used ChatGPT \cite{chatgpt} and GPT-4-turbo \cite{openai2023gpt4} to generate questions and answers, respectively.

\paragraph{Model Training} Similar to the training process implemented in Databricks, we conduct experiments across various trainers and model choices. See \Cref{tbl:more_e2e_qa_result} for an example. We note that we did not rigorously hyperparameter-tune each model and trainer combination due to the large number of experiments to perform. We believe results in \Cref{tbl:more_e2e_qa_result} may be further improved if a hyperparameter search is ran for each method.
\section{Example Model Answers}
\label{sec:example_answers}

We provide example generated answers and retrieved documents in \Cref{tbl:databricks_example_answer} and \Cref{tbl:faire_example_answer} for Databricks and Faire, respectively. 
In \Cref{tbl:databricks_example_answer}, the weaker retrievers fail to find the relevant document given the user's query. As a result, the generative models also fail to provide the correct answer.
This shows the benefit of training and using more accurate retrievers.
In \Cref{tbl:faire_example_answer}, despite all methods retrieved the correct document, the weaker generative models failed to incorporate all relevant information in its response.
This underscores the importance to fine-tune capable generative models to improve the RQA system's performance.

\section{Dataset Statistics}
\label{sec:more_dataset_statistics}

We follow the procedure outlined in \Cref{sec:more_databricks} and \Cref{sec:more_faire} to obtain $\langle q,a,p \rangle$ pairs from documentation data. For Databricks, we first use our sampling algorithm to obtain 600 passages for training, and 75 for validation and for testing.
For Faire, we sampled 300 passages for training as Faire has fewer documents, and 75 for validation and for testing.
We then use ChatGPT \cite{chatgpt} as the LLM to generate two questions for each sampled passage, and GPT-4-turbo \cite{openai2023gpt4} to generate an answer for each $\langle q,p \rangle$ pair (see \Cref{sec:Details on Data Generation} for more details).
This results in a training, evaluation, and testing dataset with statistics reported in \Cref{tbl:data_statistics}.

\section{Customizing an RQA Pipeline}
\label{sec:assembling_pipeline_full}

\framework{} uses a modular design, enabling researchers to easily assemble and customize an RQA pipeline. As an example, we provide an implementation of a (dummy) safety filter added to the \texttt{SimpleRQA} pipeline in \Cref{lst:custom_component}. In \Cref{lst:custom_component}, the \texttt{DontKnowSafetyFilter} module will ignore the answers generated by the previous components of the \texttt{SimpleRQA} modules, and always return ``I don't know.'' as an answer.

In general, users can easily add new modules to an existing pipeline by: 1) implementing a class that inherits from \texttt{Component}, which requires defining a \texttt{run} method and \texttt{run\_input\_keys}, and 2) append the module to the \texttt{components} field. Alteratively, researchers can create a fully customized pipeline by inheriting from the \texttt{RQAPipeline} class. For more documentation and examples, please refer to our GitHub pages.
\begin{table*}[t]
  \centering
  \scalebox{0.9}{
    \begin{tabular}{p{0.2\linewidth}p{0.8\linewidth}}
      \toprule
      \multicolumn{2}{c}{\textbf{Databricks RQA Example}} \\
      \midrule
      \multicolumn{2}{c}{User: How many catalogs can a metastore have in Databricks?} \\
      \midrule
      text-ada-002+\phantom{xx} GPT-3.5-turbo &
      In Databricks, a metastore can have multiple catalogs. There is \colorbox{red!30}{no specific limit on the number of catalogs} that can be associated with a metastore. Catalogs are used to organize [...] Users can see all catalogs on which they have been assigned the `USE CATALOG` data permission.\\
      text-ada-002+\phantom{xx} GPT-4-turbo &
      A metastore in Databricks can contain multiple catalogs. [...] There \colorbox{red!30}{isn't a specified limit to the number of catalogs} a metastore can have mentioned in the documentation, which suggests that users can create multiple catalogs as needed to manage their data in an organized manner [...].\\
      E5(CTL)+\phantom{xx} Mistral-7B(SFT)&
      As of \colorbox{green!30}{August 25, 2022}, a metastore can have \colorbox{green!30}{up to 1000 catalogs}.\\
      BGE(CTL)+ Starling-7B(SFT)&
      As of \colorbox{green!30}{August 25, 2022}, a metastore in Databricks can have \colorbox{green!30}{up to 1000 catalogs}.\\
      \midrule
      reference document: &
    \# <UC> GA release note
    
    **August 25, 2022**

    <UC> is now generally available on <Databricks>.
    
    **This article describes <UC> as of the date of its GA release.** It focuses primarily on the features and updates [...]
    
    \colorbox{green!30}{As of August 25, 2022}
    
    - Your <Databricks> account can have only one metastore per region
    
    - \colorbox{green!30}{A metastore can have up to 1000 catalogs.}
    
    - A catalog can have up to 10,000 schemas.
    
    - A schema can have up to 10,000 tables.
    
    For current <UC> quotas, see [...].
      \\
      \bottomrule
    \end{tabular}
  }
  \caption{Example answers generated by different RQA systems. Questions come from the test dataset for Databricks, which is described in \Cref{sec:more_dataset_statistics}. Due to space limitations, we display the most relevant document amongst all documents retrieved. In this example, text-ada-002 failed to retrieve the most relevant documents, causing GPT-3.5-turbo and GPT-4-turbo to produce incorrect answers.
  }
  \label{tbl:databricks_example_answer}
\end{table*}
\clearpage
\begin{table*}[t]
  \centering
  \scalebox{0.9}{
    \begin{tabular}{p{0.2\linewidth}p{0.8\linewidth}}
      \toprule
      \multicolumn{2}{c}{\textbf{Faire RQA Example}} \\
      \midrule
      \multicolumn{2}{c}{User: How should I crop my feature image for it to be approved?} \\
      \midrule
      text-ada-002+\phantom{xx} GPT-3.5-turbo &
      To ensure that your feature image is approved, you should crop it to a square shape. The image must \colorbox{blue!30}{fill a 1:1 ratio}, meaning that the width and height should be equal. By cropping your image to a square, you can meet this requirement and increase the chances of approval.\\
      text-ada-002+\phantom{xx} GPT-4-turbo &
      To ensure your feature image gets approved, you should follow these cropping and content guidelines:

1. **Square Crop Ratio**: Crop your image to a perfect square. The \colorbox{blue!30}{aspect ratio should be 1:1}, which means [...].

2. \colorbox{green!30}{**No Overlaid Text or Logos**}: Avoid adding any text or logos [...].

3. **Simplicity is Key**: Steer clear of using collages or images that are too busy. \colorbox{green!30}{A simple, singular focus on the product} [...].

4. **Visual Context**: \colorbox{green!30}{Provide context} by showing the product in use or in a setting where it is naturally found [...].

5. **Avoid Direct Eye Contact**: When including models, aim to \colorbox{green!30}{avoid direct eye contact} with the camera. [...]\\
      E5(CTL)+\phantom{xx} Mistral-7B(SFT)&
      To have your feature image approved, you should ensure that it is cropped to fit a square aspect ratio, which must \colorbox{blue!30}{fill a 1:1 ratio}. Additionally, it should \colorbox{green!30}{not include any words or logos}. The recommended approach is to choose a high-resolution lifestyle image that showcases your best-selling product(s) in use, such as a cheese knife with cheese and grapes on a cutting board. Also, you should \colorbox{green!30}{provide some visual context} in [...]\\
      BGE(CTL)+ Starling-7B(SFT)&
      To have your feature image approved, you should crop it to a square shape that \colorbox{blue!30}{fills a 1:1 ratio}.\\
      \midrule
      reference document: &
    \#\# Feature image guidelines
    
Crop your image to a square. It must \colorbox{blue!30}{fill a 1:1 ratio}.

\colorbox{green!30}{Don’t include overlaid text or logos}.

\colorbox{green!30}{Avoid collages and keep it simple.} Collages cause clutter in a space already crowded with many other brands. Choose a single image with a clear product focus. Simple, uncluttered often stand out more, as they provide rest to the shopper’s eye that may be weary from browsing many pages of content.

\colorbox{green!30}{Provide visual context} (avoid plain white backgrounds). While it’s true that great packaging design attracts people all by itself, showing your product in use helps get people excited about your brand and helps tell a more visual story about what you have to offer.

\colorbox{green!30}{Avoid showing models with direct eye contact.} [...]
      \\
      \bottomrule
    \end{tabular}
  }
  \caption{Example answers generated by different RQA systems. Questions come from the test dataset for Faire, which is described in \Cref{sec:more_dataset_statistics}. Due to space limitations, we display the most relevant document amongst all documents retrieved. In this example, all methods retrieved the correct document, but GPT-3.5-turbo and Starling-7B failed to include details other than ``fill a 1:1 ratio"" (c.f. \Cref{tbl:e2e_qa_result}).
  }
  \label{tbl:faire_example_answer}
\end{table*}
\begin{listing*}
\begin{minted}[mathescape,
    linenos,
    numbersep=5pt,
    gobble=3,
    framesep=2mm]{python}
    from local_rqa import ...
    ### pre-built RQA pipeline
    rqa = SimpleRQA.from_scratch(
        database_path="db_path/",
        embedding_model_name_or_path="...",
        qa_model_name_or_path="...",
    )

    ### custom module: safety filter
    class DontKnowSafetyFilter(Component):
        run_input_keys = [
            "batch_questions",
            "batch_source_documents",
            "batch_dialogue_session",
            "batch_answers",
        ]

        def run(self, *args, **kwargs):
            return RQAOutput(
                batch_answers="I don't know.",
                batch_source_documents=kwargs["batch_source_documents"],
                batch_dialogue_session=kwargs["batch_dialogue_session"],
            )
    
    rqa.components.append(DontKnowSafetyFilter())
    
    ### run QA!
    rqa.qa(...)  # output: "I don't know."
\end{minted}
\caption{Adding a custom component to the \texttt{SimpleRQA} pipeline. In general, researchers can easily extend an existing pipeline by editing the \texttt{.component} field, or create new pipelines by inherting from the \texttt{RQAPipeline} class.}
\label{lst:custom_component}
\end{listing*}
\end{document}